\documentclass{article}

\usepackage[preprint]{ARLET_2025}

\usepackage[utf8]{inputenc} 
\usepackage[T1]{fontenc}    
\usepackage{hyperref}       
\usepackage{url}            
\usepackage{booktabs}       
\usepackage{amsfonts}       
\usepackage{nicefrac}       
\usepackage{microtype}      
\usepackage{xcolor}         
\usepackage{amsmath} 
\usepackage{xcolor}
\usepackage{algorithm}
\usepackage{algpseudocode}
\usepackage{graphicx} 
\usepackage{multirow}
\usepackage{multicol}

\title{Policy Gradient Guidance Enables Test Time Control}

\author{
Jianing Qi$^{1}$,~~Hao Tang$^{1,2}$,~~Zhigang Zhu$^{1,3}$\\ 
$^1$CUNY Graduate Center, $^2$Borough of Manhattan Community College, $^3$The City College of New York\\ 
{\tt\small jqi@gradcenter.cuny.edu},
}

\begin{document}

\maketitle

\begin{abstract}
We introduce Policy Gradient Guidance (PGG), a simple extension of classifier-free guidance from diffusion models to classical policy gradient methods. PGG augments the policy gradient with an unconditional branch and interpolates conditional and unconditional branches, yielding a test-time control knob that modulates behavior without retraining. We provide a theoretical derivation showing that the additional normalization term vanishes under advantage estimation, leading to a clean guided policy gradient update. Empirically, we evaluate PGG on discrete and continuous control benchmarks. We find that conditioning dropout—central to diffusion guidance—offers gains in simple discrete tasks and low sample regimes, but dropout destabilizes continuous control. Training with modestly larger guidance ($\gamma>1$) consistently improves stability, sample efficiency, and controllability. Our results show that guidance, previously confined to diffusion policies, can be adapted to standard on-policy methods, opening new directions for controllable online reinforcement learning.\footnote{Project website: \url{user074.github.io/policy-gradient-guidance-project/}}
\end{abstract}

\section{Introduction}

Diffusion models have demonstrated success in generative modeling \cite{ho2020denoising, sohl2015deep, song2019generative, song2020score}, and one cornerstone that contributes the success is classifier-free guidance (CFG) \cite{ho2022classifierfreediffusionguidance}. CFG provides a simple yet powerful mechanism for test-time controllability. By interpolating between an unconditional and a conditional distribution, CFG steers model with more high-quality outputs without retraining. This controllability is widely adapted to nowadays' diffusion models \cite{rombach2022highresolutionimagesynthesislatent}.

Inspired by this success, recent work has explored diffusion in reinforcement learning (RL) \cite{chi2024diffusionpolicyvisuomotorpolicy, huang2023diffusion}. Diffusion-based RL methods, such as Diffuser \cite{janner2022planningdiffusionflexiblebehavior, ajay2022conditional}, utilize the CFG to condition the model towards high-reward behaviors. These methods show that CFG enables diffusion policies with improved controllability.
However, a drawback of these approaches is that they are tightly coupled to generative architectures such as diffusion or flow matching \cite{ho2020denoising, lipman2023flowmatchinggenerativemodeling}. Their effectiveness relies on the iterative sampling structure of the underlying architectures, which naturally admits the injection of guidance during generation.

In contrast, classical RL algorithms such as policy gradient methods, like Proximal Policy Optimization (PPO) \cite{schulman2017proximalpolicyoptimizationalgorithms}, use direct parametric policies that lack an iterative sampling procedure. It is very different from generative architectures. As a result, there is currently no established method for applying classifier-free guidance in standard online RL policies for test time control.

This work introduces Policy Gradient Guidance (PGG), a simple yet general mechanism for bringing classifier-free guidance into standard policy gradient methods. Our approach augments a PPO policy with an unconditional branch and defines a $\gamma$ weighted mixture of conditional and unconditional logits similar as CFG in diffusion. This enables controllable inference in PPO without architectural dependence on diffusion or flow models.

Our contributions are the following:
\begin{itemize}
    \item \textbf{Classifier-Free Guidance in PPO:} We propose the first extension of classifier-free guidance to classical policy gradient methods, introducing \emph{PGG} as a simple and general framework for inference-time controllability.
    \item \textbf{Inference-Time Control Knob:} We demonstrate that guidance acts as a tunable knob at test time, allowing modulation of policy behavior without retraining.
    \item \textbf{Training Strategies:} Through empirical study, we show that conditioning dropout is effective in simple environments, while introducing a biased guidance parameter $\gamma \!>\! 1$ during training yields stronger and more stable performance in complex tasks.
    \item \textbf{Bridging Diffusion and Classical RL:} Our work highlights that controllability mechanisms previously restricted to offline diffusion-style policies can be adapted to standard on-policy RL, broadening the applicability of guidance beyond generative architectures.
\end{itemize}

\section{Related Work}

\paragraph{Policy Gradient}
Our method builds on on-policy reinforcement learning, where policy gradient algorithms such as REINFORCE \cite{sutton1999policy}, TRPO \cite{schulman2015trust}, and PPO \cite{schulman2017proximalpolicyoptimizationalgorithms} remain the standard approaches. These methods directly optimize a parametric policy with respect to long-term returns, often augmented with critic baselines and advantage estimation \cite{schulman2018highdimensionalcontinuouscontrolusing}. 
Significant amount of the works are built on the policy gradient methods and it is one of the widely adapted methods in RL \cite{ouyang2022traininglanguagemodelsfollow, deepseekai2025deepseekr1incentivizingreasoningcapability, openai2019dota2largescale}. 
Extend from this we provide a test time control knob for the policy gradient methods.

\paragraph{Guidance in Generative RL}
Classifier-free guidance (CFG) was first proposed in diffusion models \cite{ho2022classifierfreediffusionguidance}, where interpolating conditional and unconditional predictions yields controllable generation without a classifier. Inspired by this, guidance has been widely explored in diffusion-based RL: planning with diffusion \cite{janner2022planningdiffusionflexiblebehavior}, HDMI \cite{pmlr-v202-li23ad}, Decision Diffuser \cite{ajay2022conditional}, and CFGRL \cite{frans2025diffusionguidancecontrollablepolicy}.

All of the above approaches remain tied to offline diffusion or flow architectures. In contrast, our work explores \emph{whether classifier-free guidance can be applied to standard policy gradient methods such as PPO}, providing controllability at inference time without requiring diffusion or flow architectures.

\section{Background}
\label{sec:cfg}

\paragraph{Classifier-Free Guidance.}

In generative modeling, an interesting property is to balance the diversity and the quality with conditional control \cite{brock2018large, dhariwal2021diffusion}. 
Classifier-Free Guidance (CFG) emerged as a simple but powerful solution in diffusion models \cite{ho2022classifierfreediffusionguidance}. They key goal for the diffusion is to estimate the \textbf{score function} $\nabla_x \log p(x|y)$ given input $x$ and condition $y$, and CFG gives additional control to improve the quality with a $\gamma$ term.

At inference time, CFG introduces a single parameter $\gamma$ that smoothly interpolates between two distributions: 
purely unconditional generation ($\gamma=0$) and fully conditional generation ($\gamma=1$). 
Intermediate values amplify the influence of the condition while still respecting the data distribution, but the most interesting part is when $\gamma > 1$ whereas the quality can improve even further.

The key idea is to interpolate between unconditional and conditional score functions: given input $x$ and condition $y$, the guided score is
\begin{equation}
\nabla_x \log \hat{p}(x|y) = (1-\gamma)\nabla_x \log p(x) + \gamma \nabla_x \log p(x|y),
\end{equation}
where $\gamma$ is the guidance strength.  

Equivalently, by reverting the gradient and the log, the reweighted distribution is
\begin{equation}
\hat{p}(x|y) \propto p(x)^{1-\gamma} \, p(x|y)^{\gamma},
\label{eq:gudiance_distribution}
\end{equation}
showing that CFG is effectively a multiplicative interpolation in probability space \cite{dieleman2022guidance}.  

\paragraph{Policy Gradient.}
In reinforcement learning, the policy gradient theorem expresses the gradient of the expected return with respect to the parameter $\theta$ as a \textbf{score function} too. The gradient of the expected return with respect to the policy parameter $\theta$ is
\begin{equation}
\nabla_\theta J(\theta) = \mathbb{E}\!\left[ A \, \nabla_\theta \log \pi_\theta(a|s)\right],
\end{equation}
where $A$ is the advantage function $A= Q(s,a) - V(s)$.   
This formulation arises from the likelihood-ratio trick and establishes the score function $\nabla_\theta \log \pi_\theta(a|s)$ as the key update direction.  

\paragraph{Motivation: Guidance for Policies.}
The analogy between CFG and policy gradient raises a compelling question: \textit{can we introduce a guidance parameter $\gamma$ to RL allowing test-time control of policies similar to CFG in diffusion?}

If feasible, this would enable controllable deployment of policies without retraining, similar as CFG in diffusion models. Moreover, it might improve the exploitation of the policy during test time. 

However, a direct derivation from Baye's rule following the CFG fails because policy gradients differentiate with respect to the parameter $\theta$ rather than the input $x$. 
We show such derivation problem from Bayes’ rule is deferred to Appendix~\ref{appendix:cfg}.
This introduces intractable terms involving the marginal state distribution $\pi_\theta(s)$, which depends on both the policy and environment dynamics.  
While recent work circumvents this issue in flow-matching frameworks for offline RL \cite{frans2025diffusionguidancecontrollablepolicy}, widely used methods such as PPO remain parameter-gradient–based.

This motivates our study: extending guidance principles to standard policy-gradient algorithms to enable controllable test-time behavior.

\section{Method}
\label{sec:method}

\paragraph{From Guidance in Diffusion to Guidance in Policies.}
Recall from previous Section~\ref{sec:cfg} Equation \ref{eq:gudiance_distribution} that Classifier-Free Guidance (CFG) constructs a new distribution by interpolating between unconditional and conditional score functions. 
This can be viewed as composing a new powered product interpolation in the probability space, balancing the base distribution $p(x)$ and the conditional $p(x|y)$. 

We propose an analogous construction for policies. 
Instead of interpolating distributions over the data $x$, we interpolate distributions over the action $a$:

\begin{equation}
\boxed{\;\hat{\pi}_\theta(a|s)\;\propto\;\pi_\theta(a)^{\,1-\gamma}\,\pi_\theta(a|s)^{\,\gamma}\;}
\end{equation}

Here, $\pi_\theta(a)$ is the \textit{unconditional policy} over the action, independent of the state $s$, $\pi_\theta(a|s)$ is the \textit{conditional policy}, the standard actor conditioned on the state $s$, $\gamma$ is the guidance strength: $\gamma=0$ recovers unconditional behavior, $\gamma=1$ recovers the standard policy, and $\gamma>1$ extrapolates beyond the standard policy.  

This mirrors the CFG formulation, but applied in the policy space directly. We call it \textit{Policy Gradient Guidance (PGG)}. 

\paragraph{Derivatives with respect to actions.}
To see the analogy more clearly, consider taking the log and differentiating with respect to actions $a$. 
The normalization terms vanish, yielding the standard CFG-like form:
\begin{equation}
\nabla_a \log \hat{\pi}_\theta(a|s) 
= \gamma \nabla_a \log \pi_\theta(a|s) + (1-\gamma)\nabla_a \log \pi_\theta(a).
\label{eq:diffusion_style_derivative}
\end{equation}
This highlights that the guidance interpolates between unconditional and conditional influences on actions. If we have a diffusion model architecture we can apply it directly. However, adapting existing networks into diffusion model might take a lot of effort and substantial design changes.

\paragraph{Derivatives with respect to parameters.}
The key difference emerges when differentiating with respect to the policy parameter $\theta$. 
Let $Z_\theta(s)$ denote the normalization constant ensuring $\hat{\pi}_\theta(\cdot|s)$ sums to one. 
Then,
\begin{equation}
\nabla_\theta \log \hat{\pi}_\theta(a|s) 
= \gamma \nabla_\theta \log \pi_\theta(a|s) 
+ (1-\gamma) \nabla_\theta \log \pi_\theta(a) 
- \nabla_\theta \log Z_\theta(s).
\label{eq:direct_derivative}
\end{equation}
Unlike the action-derivative case in Equation \ref{eq:diffusion_style_derivative}, here the state-dependent term $\nabla_\theta \log Z_\theta(s)$ does not vanish automatically.

\paragraph{Policy Gradient Guidance.}
However, we can get some interesting results by plugging Equation \ref{eq:direct_derivative} into the policy gradient theorem. We can pull out the $Z$ term since it is state dependent not action dependent.
\begin{equation*}
    \begin{aligned}
    \nabla_\theta J(\theta) 
    &= \mathbb{E}\!\left[A \, \nabla_\theta \log \hat{\pi}_\theta(a|s)\right] \\
    &= \mathbb{E}\!\left[A \, \big(\gamma \nabla_\theta \log \pi_\theta(a|s) 
    + (1-\gamma)\nabla_\theta \log \pi_\theta(a)\big)\right] 
    - \mathbb{E}_{s \sim \hat{\pi}}\!\left[\nabla_\theta \log Z_\theta(s) \, \mathbb{E}_{a \sim \hat{\pi}}[A]\right].
    \end{aligned}
\end{equation*}

We compute advantages from rollouts generated by the guided policy $\hat{\pi}$. Now, recall that $\mathbb{E}_{a \sim \hat{\pi}}[A]=0$ by definition of the advantage function:  
\[
\mathbb{E}_{a \sim \hat{\pi}}[A] 
= \mathbb{E}_{a \sim \hat{\pi}}[Q(s,a) - V(s)] 
= \mathbb{E}_{a \sim \hat{\pi}}[Q(s,a)] - V(s) = 0.
\]  
Thus, the state-only normalization term vanishes under expectation.  

The resulting guided policy gradient is therefore
\begin{equation}
\nabla_\theta J(\theta) 
= \mathbb{E}\!\left[A \, \big(\gamma \nabla_\theta \log \pi_\theta(a|s) 
+ (1-\gamma)\nabla_\theta \log \pi_\theta(a)\big)\right].
\end{equation}

This shows that, when integrated with advantage estimation, our guidance formulation behaves analogously to CFG: 
the gradient is an interpolation between conditional and unconditional updates, controlled by $\gamma$. 
For $\gamma=1$, the update reduces to the standard policy gradient; for $\gamma\neq 1$, we obtain controllable guidance of policies at both training and test time.

\section{Experiments}
\paragraph{Setup.}
We evaluate our proposed guidance formulation on both continuous- and discrete-action benchmarks. 
Our implementation builds directly on the PPO reference from CleanRL \cite{huang2022cleanrl}, with minimal modifications to incorporate guidance (Algorithm~\ref{alg:cfg-ppo}). 
Specifically, we extend the actor network with a learnable \emph{null embedding}, which serves as the unconditional branch $\pi_\theta(a|\emptyset)$. 
This enables us to construct the guided policy
\[
\hat{\pi}_\theta(a|s) \propto \pi_\theta(a)^{1-\gamma}\,\pi_\theta(a|s)^\gamma,
\]
while preserving the standard PPO optimization loop.

\paragraph{Conditioning dropout.}
A key open question is whether conditioning dropout—central to classifier-free guidance in diffusion models—remains effective in reinforcement learning. 
Unlike static generative settings, the environment dynamics in RL depend directly on the policy distribution, so dropout could interact with training stability in unexpected ways. 
We therefore include ablations with and without dropout, toggling whether the conditional branch is replaced by the null embedding during policy updates in Section \ref{sec:condition} and Section \ref{sec:gamma-no-dropout}.

\paragraph{Environments.}
We evaluate across both continuous- and discrete-action tasks:  
\begin{itemize}
    \item \textbf{Continuous:} MuJoCo v4 \cite{todorov2012mujoco} environments HalfCheetah, Hopper, Humanoid, Inverted Pendulum, Pusher, and Walker2d.  
    \item \textbf{Discrete:} CartPole-V1 and Acrobot-V1 \cite{brockman2016openai}.
\end{itemize}
These environments cover diverse dynamics and action spaces, allowing us to probe the generality of guidance in PPO.

\paragraph{Evaluation protocol.}
For each environment, we train 5 models with different random seeds. 
Each trained policy is then evaluated over 50 test episodes, reporting the mean and standard deviation of episodic returns. We also save intermediate checkpoints for each model.
We compare against the \textbf{vanilla PPO baseline} (both discrete and continuous versions) implemented in CleanRL. 

\paragraph{Implementation details.}
Unless otherwise noted, we follow the default PPO hyperparameters from CleanRL.
Guidance introduces new hyperparameters: the guidance strength $\gamma$ (default $1.0$, i.e.\ standard PPO), and a dropout probability $p_{\text{drop}}$ (default $0.1$ during updates).
Training is run for $5\times10^5$ timesteps for discrete and $1\times10^6$ timesteps for continuous. In our implementation $\hat{\pi}$ is parameterized directly as a normalized distribution. For discrete actions we use a softmax over guided logits, and for continuous actions we use a Gaussian with guided mean and shared covariance.

\begin{algorithm}[t]
\caption{PPO with Policy Gradient Guidance (PGG)}
\label{alg:cfg-ppo}
\small
\begin{algorithmic}[1]
\State Initialize policy $\pi_\theta(a|s)$, value function $V_\psi(s)$
\State Add unconditional branch $\pi_\theta(a|\emptyset)$ via learnable null embedding \Comment{Add one parameter}
\For{each iteration}
  \State Collect rollouts $\{s_t,a_t,r_t\}$ using guided policy   \Comment{The only change in PPO}
  \[
  \hat{\pi}_\theta(a|s) \propto \pi_\theta(a|\emptyset)^{1-\gamma}\,\pi_\theta(a|s)^\gamma
  \]
  \State Compute advantages $\{A_t\}$ (e.g.\ GAE)
  \State Update $\theta,\psi$ with PPO clipped objective, but using $\hat{\pi}_\theta$ in place of $\pi_\theta$
\EndFor
\end{algorithmic}
\end{algorithm}

\section{Conditional Dropout}
\label{sec:condition}
Classifier-Free Guidance in diffusion models often relies on \emph{random dropout} to encourage the model to learn both conditional and unconditional branches robustly. 
The intuition is that by occasionally dropping the conditional input, the unconditional pathway receives sufficient training signal. 
In reinforcement learning, however, the setting is more delicate: the policy determines the distribution of visited states, and perturbations from dropout may alter the policy and environment in unexpected ways. 
We therefore evaluate whether conditioning dropout remains beneficial when applied to PPO.

\paragraph{Discrete control.}
On CartPole and Acrobot, we compare PPO with guided variants that apply $10\%$ random dropout to the conditional branch during updates. 

\begin{figure}
    \centering
    \includegraphics[width=0.9\linewidth]{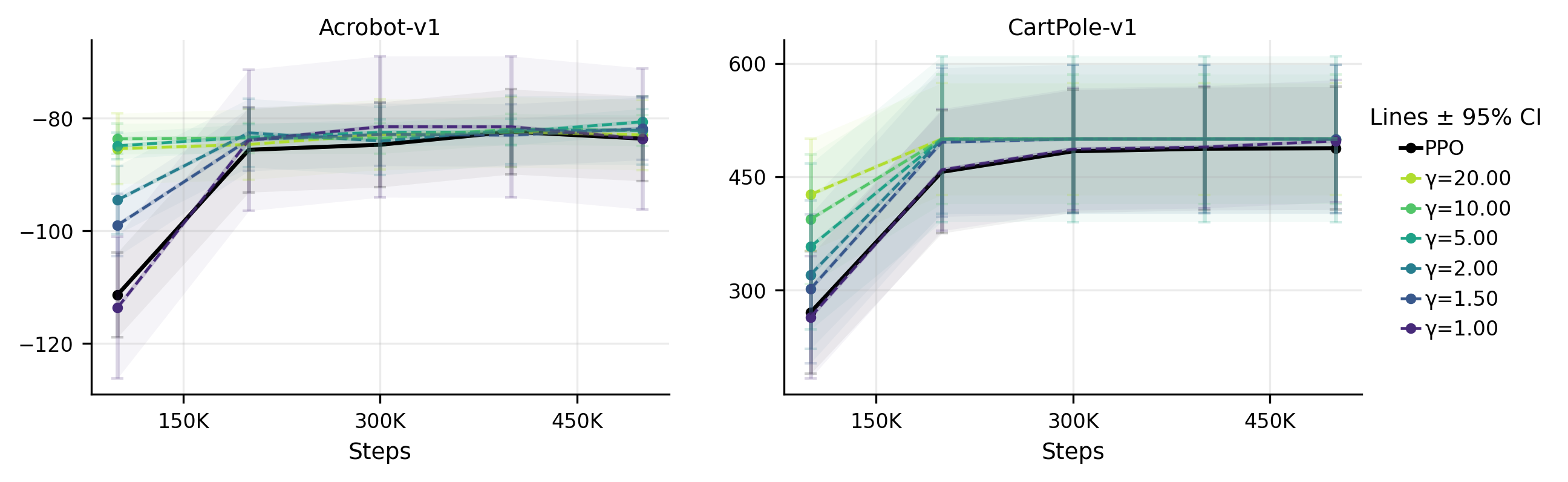}
    \caption{Discrete PPO with Dropout 10\%. We compare the performance across training steps vs the reward, and we increase the guidance strength of the model from $\gamma=1$ to $\gamma=20$ during inference.}
    \label{fig:discrete_dropout}
\end{figure}

As shown in Figure \ref{fig:discrete_dropout}, guidance with dropout substantially accelerates inference performance compared to vanilla PPO. 
When $\gamma=1$, performance is similar to PPO, but increasing $\gamma$ at test time—without retraining—significantly boosts returns, particularly in the early stages (e.g., 100k steps).

\paragraph{Continuous control.}
For MuJoCo benchmarks (HalfCheetah, Hopper, Walker2d, Humanoid, InvertedPendulum, Pusher), results are more mixed. 
\begin{figure}
    \centering
    \includegraphics[width=0.9\linewidth]{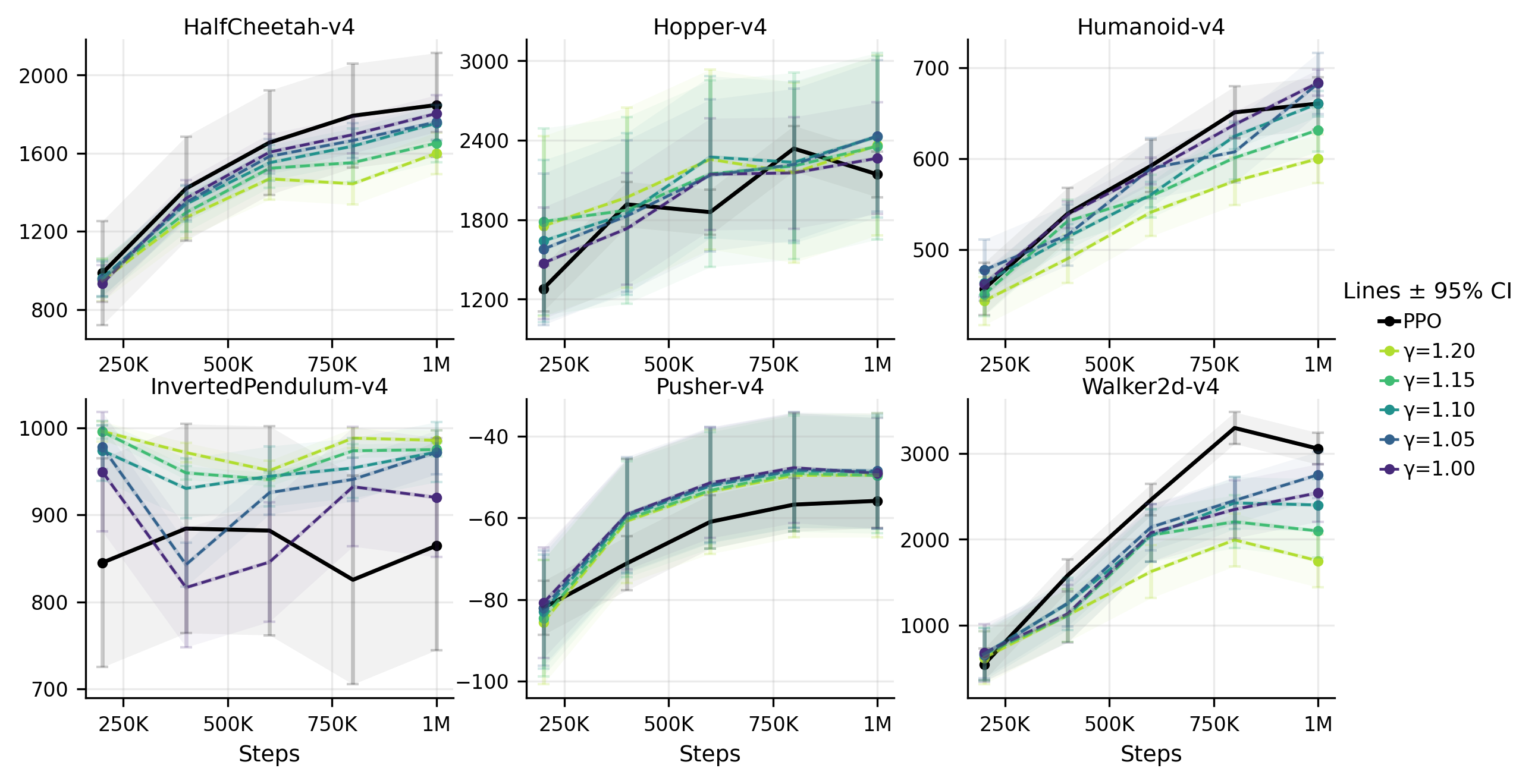}
    \caption{Continuous PPO with Dropout 10\%. $\gamma$ value increases as represented from blue to green. We can see dropout resulted mixed results, and increase $\gamma$ might hurt the performance.}
    \label{fig:continuous_dropout}
\end{figure}

Unlike the discrete case, where performance improves monotonically with larger $\gamma$, in continuous control too large a deviation from the learned policy distribution can degrade performance. 
Most environments achieve their best results with mild or no guidance ($\gamma \in [1.0, 1.05]$), while larger $\gamma$ values introduce instability, especially in high dimensional tasks such as HalfCheetah, Humanoid, and Walker2d. The performance decreases with larger $\gamma$.

\paragraph{Analysis.}
These results suggest that conditional dropout in PPO behaves differently from its role in diffusion models. 
In discrete environments, dropout reliably improves sample efficiency, likely because the state distribution is simple and perturbations do not destabilize training. 
In continuous environments, however, dropout provides little or no benefit, and can even harm asymptotic performance. 
In practice, the best results often occur at $\gamma=1.0$, which corresponds to no effective contribution from the unconditional branch. 

This contrast suggests that instead of relying on stochastic dropout to train the unconditional branch, a more principled and stable approach is to directly adjust the guidance strength $\gamma>1$, which naturally integrates into the policy gradient update (Section~\ref{sec:gamma-no-dropout}).

\section{Replacing Dropout with Larger Guidance During Training}
\label{sec:gamma-no-dropout}

\paragraph{Motivation.}
While conditioning dropout can encourage learning in the unconditional branch, it also perturbs on-policy rollouts and may destabilize training. 
Our guided policy gradient (Section~\ref{sec:method}) suggests a simpler alternative: train \emph{without} dropout ($p_{\text{drop}}=0$) and set $\gamma>1$, so that

\[
\nabla_\theta J(\theta)
= \mathbb{E}\!\left[A \,\big(\gamma \nabla_\theta \log \pi_\theta(a|s) + (1-\gamma)\nabla_\theta \log \pi_\theta(a)\big)\right],
\]
where $(1-\gamma)<0$ for $\gamma>1$. 
Intuitively, the unconditional branch is trained to model \emph{what not to do} (negative weight on high-advantage actions), while the conditional branch is amplified, yielding an alternative to random dropout. It ensures that both branches receive a gradient signal throughout training.  

We use the same setup of CleanRL PPO backbone with our unconditional branch via a learnable null embedding and guided policy
$\hat{\pi}_\theta(a|s)\propto \pi_\theta(a|\emptyset)^{1-\gamma}\pi_\theta(a|s)^\gamma$.
For training we use $\gamma=1.1$ to mimic the 10\% updates to the unconditional part.
No dropout is used anywhere.
During evaluation, we sweep $\gamma$ to study how test-time controllability interacts with this training setup.

\begin{figure}
    \centering
    \includegraphics[width=0.9\linewidth]{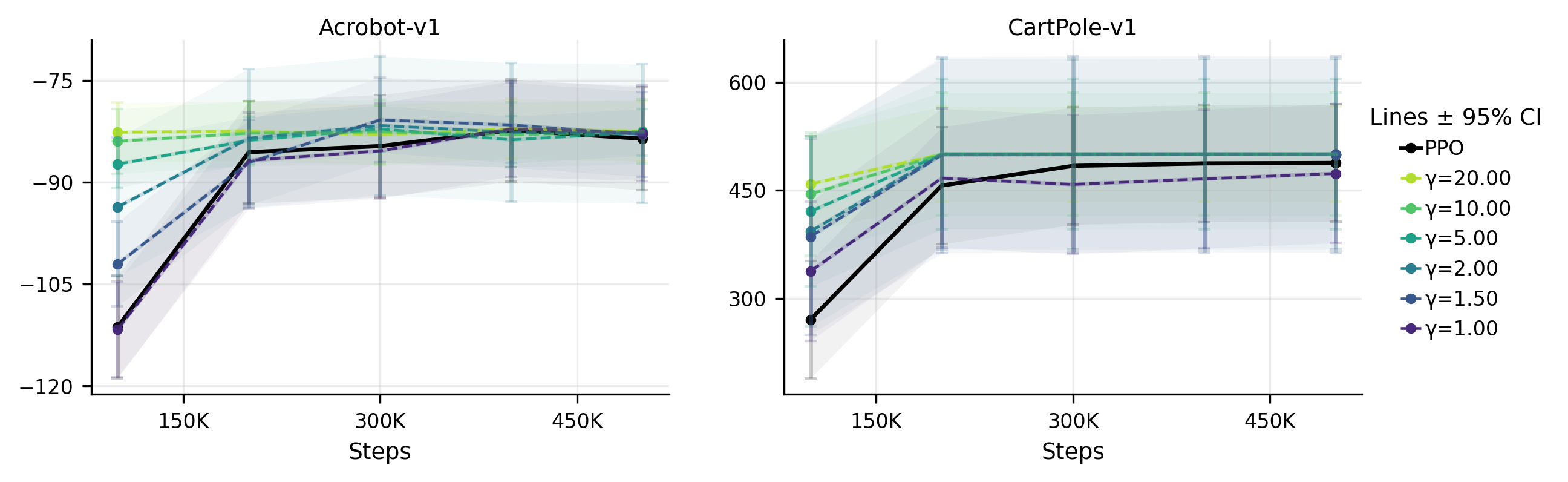}
    \caption{Discrete PPO with training $\gamma=1.1$}
    \label{fig:discrete_gamma}
\end{figure}

\subsection*{Discrete control}
As shown in Figure~\ref{fig:discrete_gamma}, performance improves monotonically with $\gamma$, echoing the earlier dropout-based experiments. 
Compared to vanilla PPO, training with $\gamma=1.1$ yields faster convergence and more stable returns across CartPole and Acrobot. Detailed results are in Appendix \ref{sec:detailed_results} Table \ref{tab:discrete_results}.

\subsection*{Continuous control}

The more informative case is continuous control, where dropout alone had little effect. 
Here, training with $\gamma=1.1$ produces markedly different dynamics. 
At evaluation, we sweep $\gamma \in \{1.0, 1.05, 1.1, 1.15, 1.2, 1.25, 1.3\}$ to probe robustness. 
Detailed results are in Appendix \ref{sec:detailed_results} Table \ref{tab:continuous_results}. 

\begin{figure}
    \centering
    \includegraphics[width=0.9\linewidth]{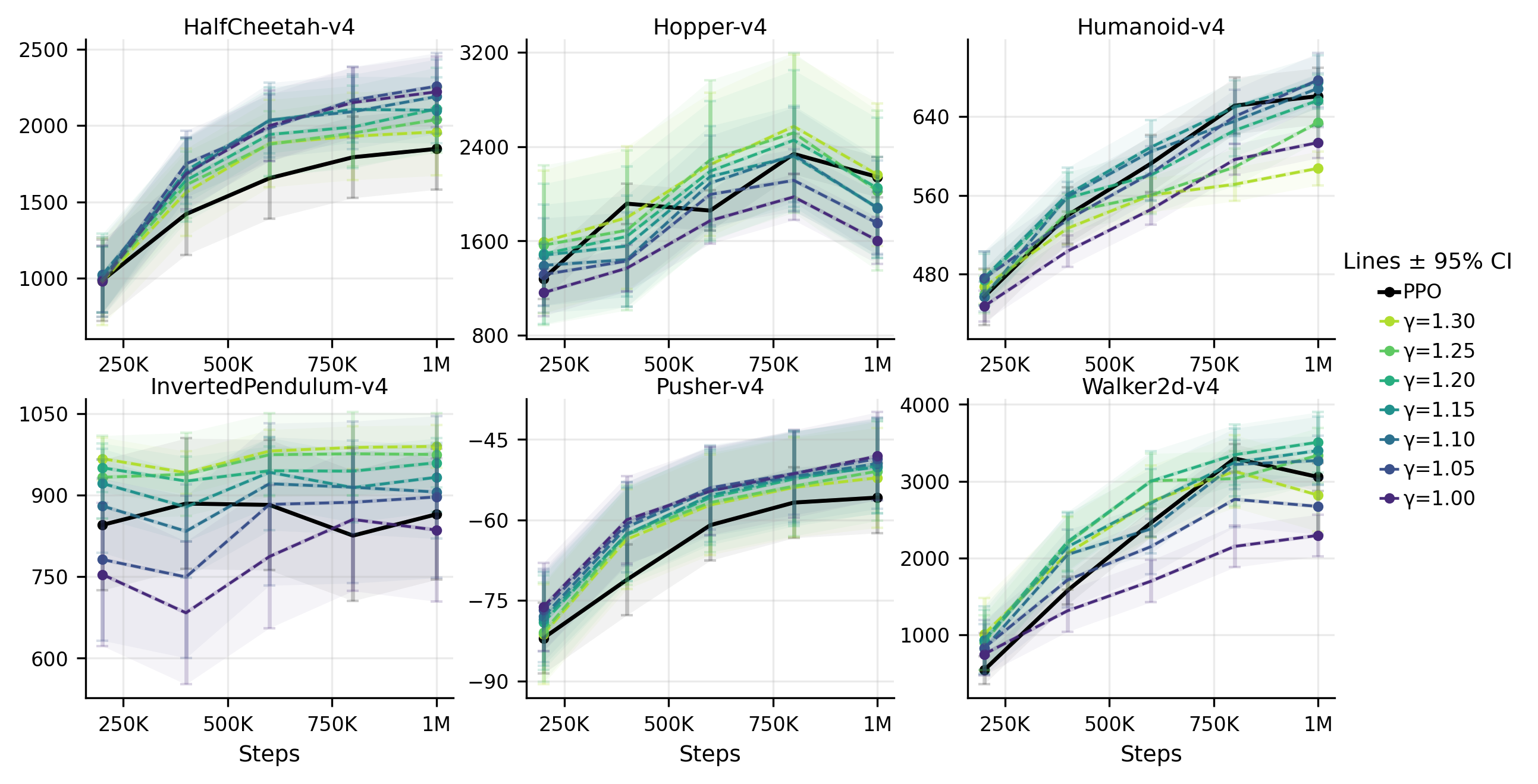}
    \caption{Continuous PPO with training $\gamma=1.1$. $\gamma$ value increases as represented from blue to green. We can see increase $\gamma$ can improve performance in most cases until $\gamma=1.3$, which there is a sharp drop in more complex cases. In all cases, an optimal $\gamma$ can outperform PPO..}
    \label{fig:continuous_gamma}
\end{figure}

The results are summarized in Figure~\ref{fig:continuous_gamma}. They show that:
\begin{itemize}
    \item \textbf{HalfCheetah, Pusher:} guided policies surpass PPO baselines; larger $\gamma$ improves early returns but too large values destabilize training.  
    \item \textbf{Hopper, InvertedPendulum:} increasing $\gamma$ consistently improves performance and exceeds PPO.  
    \item \textbf{Humanoid, Walker2d:} moderate guidance ($\gamma \approx 1.1$--$1.2$) yields the best results; larger $\gamma$ degrades asymptotic returns.  
\end{itemize}
Overall we do see a much signficant effect of the $\gamma$-increase and performance-increase relationships in the continouos case when we train it with a bit higher $\gamma$ compared to drop out.

\paragraph{Analysis}
Replacing dropout with modestly larger $\gamma$ during training leads to a more reliable unconditional branch and better alignment with the guided-gradient view. 
This approach improves sample efficiency, supports controllable test-time behavior, and avoids the instability observed with dropout. 
Overall, $\gamma$ extrapolation emerges as a more stable and effective mechanism, especially in continuous control tasks, provided that $\gamma$ is not set excessively large.

\section{Discussion}

\subsection{Guidance Derivation}
In our formulation, we followed the analogy $\hat{p}(x|y) \propto p(x)^{1-\gamma} p(x|y)^\gamma$, treating the action $a$ as $x$ and the state $s$ as $y$. 
This is the most natural choice in online policy optimization, but it is not the only possibility. 
For example, guidance could be conditioned on \emph{goals} or auxiliary signals, as explored in offline RL settings \cite{frans2025diffusionguidancecontrollablepolicy}. 

The key challenge is stability: our derivation relies on the cancellation of the $Z(s)$ term, which in turn depends on the advantage estimator being unbiased. 
A biased or none-converged advantage might introduce the normalization term and cause instability. Thus it is crucial to ensure the critic to be well trained.
Alternative conditions may require rethinking how the advantage function is defined and how the normalizer behaves.

Another question is: What does $\pi_\theta(a|\emptyset)$ mean? In our interpretation, it represents an unconditional action prior. It is the distribution over actions when state information is unavailable or ignored. This can be interpreted as the 'default' action tendencies learned across all states encountered during training.

\subsection{Effects of $\gamma$}
The choice of $\gamma$ is critical. 
In diffusion models, $\gamma$ can often be pushed to large values without collapse, amplifying the conditional signal. 
In reinforcement learning, however, the picture is more nuanced:
In simple tasks (e.g., CartPole, Acrobot), $\gamma$ can be made quite large and still improves performance monotonically.  
In complex tasks (e.g., Hopper, Humanoid, Walker2d), stability deteriorates quickly once $\gamma$ exceeds a modest range, and performance drops sharply.  

This suggests that $\gamma$ plays a dual role: as a knob for test-time controllability and as a regularization strength during training. 
Practical use therefore requires careful tuning or even dynamic scheduling of $\gamma$.

\subsection{Limitations}
While guidance improves performance in many cases, several limitations remain. The unconditional policy ($\gamma\neq 1$) can be weaker than vanilla PPO in some environments such as Hopper, Humanoid, and Walker2d, even though extrapolated versions with $\gamma>1$ can match or outperform PPO. 

Unlike dropout, which adds negligible overhead, using $\gamma>1$ requires evaluating both conditional and unconditional branches at each update. 
Calculating a separate unconditional policy without drop out roughly doubles the compute compared to vanilla PPO.  
    
Despite these caveats, guidance provides a promising mechanism for controllable test-time behavior and suggests new directions for integrating conditional signals into online policy gradient methods.

\section{Conclusion}
We introduced Policy Gradient Guidance (PGG), the first extension of classifier-free guidance to classical on-policy reinforcement learning. By augmenting PPO with an unconditional branch and interpolating conditional and unconditional logits, PGG enables a simple test-time control knob analogous to diffusion guidance. Our theoretical analysis shows that the additional normalization term vanishes under advantage estimation, yielding a clean policy gradient update.

Empirically, we find that guidance provides controllability without retraining and can improve sample efficiency. Conditioning dropout, while effective in discrete control, proves unstable in continuous domains; replacing dropout with biased $\gamma>1$ training offers a more reliable mechanism, producing stronger returns and more robust controllability.

Taken together, our results highlight that the diffusion models' guidance can be adapted to standard policy gradient methods, opening up new avenues for controllable online reinforcement learning. Future work includes extending PGG to richer conditioning signals (e.g., goals, preferences), developing adaptive or scheduled $\gamma$ strategies to mitigate instability in high-dimensional tasks, and training on a better unconditional policy such as behavior cloning.


\subsubsection*{Acknowledgments}
\label{sec:acknowledgements}
The work is supported by the National Science Foundation (NSF) through Awards \#2131186 (CISE-MSI),  \#1827505 (PFI), and the US Air Force Office of Scientific Research (AFOSR) via Award \#FA9550-21-1-0082. The work is also supported by a College-wide Research Vision (CRV) Fund from the CCNY Provost's Office and the Google CyberNYC Initiative. This work used Google Cloud through the CloudBank project, which is supported by National Science Foundation Award \#1925001.
\bibliographystyle{abbrv}
\bibliography{citation}

\begin{thebibliography}{10}

\bibitem{ajay2022conditional}
A.~Ajay, Y.~Du, A.~Gupta, J.~Tenenbaum, T.~Jaakkola, and P.~Agrawal.
\newblock Is conditional generative modeling all you need for decision-making?
\newblock {\em arXiv preprint arXiv:2211.15657}, 2022.

\bibitem{brock2018large}
A.~Brock, J.~Donahue, and K.~Simonyan.
\newblock Large scale gan training for high fidelity natural image synthesis.
\newblock {\em arXiv preprint arXiv:1809.11096}, 2018.

\bibitem{brockman2016openai}
G.~Brockman, V.~Cheung, L.~Pettersson, J.~Schneider, J.~Schulman, J.~Tang, and W.~Zaremba.
\newblock Openai gym.
\newblock {\em arXiv preprint arXiv:1606.01540}, 2016.

\bibitem{chi2024diffusionpolicyvisuomotorpolicy}
C.~Chi, Z.~Xu, S.~Feng, E.~Cousineau, Y.~Du, B.~Burchfiel, R.~Tedrake, and S.~Song.
\newblock Diffusion policy: Visuomotor policy learning via action diffusion, 2024.

\bibitem{deepseekai2025deepseekr1incentivizingreasoningcapability}
DeepSeek-AI, D.~Guo, D.~Yang, H.~Zhang, J.~Song, R.~Zhang, R.~Xu, Q.~Zhu, S.~Ma, P.~Wang, X.~Bi, X.~Zhang, X.~Yu, Y.~Wu, Z.~F. Wu, Z.~Gou, Z.~Shao, Z.~Li, Z.~Gao, A.~Liu, B.~Xue, B.~Wang, B.~Wu, B.~Feng, C.~Lu, C.~Zhao, C.~Deng, C.~Zhang, C.~Ruan, D.~Dai, D.~Chen, D.~Ji, E.~Li, F.~Lin, F.~Dai, F.~Luo, G.~Hao, G.~Chen, G.~Li, H.~Zhang, H.~Bao, H.~Xu, H.~Wang, H.~Ding, H.~Xin, H.~Gao, H.~Qu, H.~Li, J.~Guo, J.~Li, J.~Wang, J.~Chen, J.~Yuan, J.~Qiu, J.~Li, J.~L. Cai, J.~Ni, J.~Liang, J.~Chen, K.~Dong, K.~Hu, K.~Gao, K.~Guan, K.~Huang, K.~Yu, L.~Wang, L.~Zhang, L.~Zhao, L.~Wang, L.~Zhang, L.~Xu, L.~Xia, M.~Zhang, M.~Zhang, M.~Tang, M.~Li, M.~Wang, M.~Li, N.~Tian, P.~Huang, P.~Zhang, Q.~Wang, Q.~Chen, Q.~Du, R.~Ge, R.~Zhang, R.~Pan, R.~Wang, R.~J. Chen, R.~L. Jin, R.~Chen, S.~Lu, S.~Zhou, S.~Chen, S.~Ye, S.~Wang, S.~Yu, S.~Zhou, S.~Pan, S.~S. Li, S.~Zhou, S.~Wu, S.~Ye, T.~Yun, T.~Pei, T.~Sun, T.~Wang, W.~Zeng, W.~Zhao, W.~Liu, W.~Liang, W.~Gao, W.~Yu, W.~Zhang, W.~L. Xiao, W.~An, X.~Liu, X.~Wang, X.~Chen, X.~Nie,
  X.~Cheng, X.~Liu, X.~Xie, X.~Liu, X.~Yang, X.~Li, X.~Su, X.~Lin, X.~Q. Li, X.~Jin, X.~Shen, X.~Chen, X.~Sun, X.~Wang, X.~Song, X.~Zhou, X.~Wang, X.~Shan, Y.~K. Li, Y.~Q. Wang, Y.~X. Wei, Y.~Zhang, Y.~Xu, Y.~Li, Y.~Zhao, Y.~Sun, Y.~Wang, Y.~Yu, Y.~Zhang, Y.~Shi, Y.~Xiong, Y.~He, Y.~Piao, Y.~Wang, Y.~Tan, Y.~Ma, Y.~Liu, Y.~Guo, Y.~Ou, Y.~Wang, Y.~Gong, Y.~Zou, Y.~He, Y.~Xiong, Y.~Luo, Y.~You, Y.~Liu, Y.~Zhou, Y.~X. Zhu, Y.~Xu, Y.~Huang, Y.~Li, Y.~Zheng, Y.~Zhu, Y.~Ma, Y.~Tang, Y.~Zha, Y.~Yan, Z.~Z. Ren, Z.~Ren, Z.~Sha, Z.~Fu, Z.~Xu, Z.~Xie, Z.~Zhang, Z.~Hao, Z.~Ma, Z.~Yan, Z.~Wu, Z.~Gu, Z.~Zhu, Z.~Liu, Z.~Li, Z.~Xie, Z.~Song, Z.~Pan, Z.~Huang, Z.~Xu, Z.~Zhang, and Z.~Zhang.
\newblock Deepseek-r1: Incentivizing reasoning capability in llms via reinforcement learning, 2025.

\bibitem{dhariwal2021diffusion}
P.~Dhariwal and A.~Nichol.
\newblock Diffusion models beat gans on image synthesis.
\newblock {\em Advances in neural information processing systems}, 34:8780--8794, 2021.

\bibitem{dieleman2022guidance}
S.~Dieleman.
\newblock Guidance: a cheat code for diffusion models, 2022.

\bibitem{frans2025diffusionguidancecontrollablepolicy}
K.~Frans, S.~Park, P.~Abbeel, and S.~Levine.
\newblock Diffusion guidance is a controllable policy improvement operator, 2025.

\bibitem{ho2020denoising}
J.~Ho, A.~Jain, and P.~Abbeel.
\newblock Denoising diffusion probabilistic models.
\newblock {\em Advances in neural information processing systems}, 33:6840--6851, 2020.

\bibitem{ho2022classifierfreediffusionguidance}
J.~Ho and T.~Salimans.
\newblock Classifier-free diffusion guidance, 2022.

\bibitem{huang2022cleanrl}
S.~Huang, R.~F.~J. Dossa, C.~Ye, J.~Braga, D.~Chakraborty, K.~Mehta, and J.~G. Araújo.
\newblock Cleanrl: High-quality single-file implementations of deep reinforcement learning algorithms.
\newblock {\em Journal of Machine Learning Research}, 23(274):1--18, 2022.

\bibitem{huang2023diffusion}
S.~Huang, Z.~Wang, P.~Li, B.~Jia, T.~Liu, Y.~Zhu, W.~Liang, and S.-C. Zhu.
\newblock Diffusion-based generation, optimization, and planning in 3d scenes.
\newblock In {\em Proceedings of the IEEE/CVF Conference on Computer Vision and Pattern Recognition}, pages 16750--16761, 2023.

\bibitem{janner2022planningdiffusionflexiblebehavior}
M.~Janner, Y.~Du, J.~B. Tenenbaum, and S.~Levine.
\newblock Planning with diffusion for flexible behavior synthesis, 2022.

\bibitem{pmlr-v202-li23ad}
W.~Li, X.~Wang, B.~Jin, and H.~Zha.
\newblock Hierarchical diffusion for offline decision making.
\newblock In A.~Krause, E.~Brunskill, K.~Cho, B.~Engelhardt, S.~Sabato, and J.~Scarlett, editors, {\em Proceedings of the 40th International Conference on Machine Learning}, volume 202 of {\em Proceedings of Machine Learning Research}, pages 20035--20064. PMLR, 23--29 Jul 2023.

\bibitem{lipman2023flowmatchinggenerativemodeling}
Y.~Lipman, R.~T.~Q. Chen, H.~Ben-Hamu, M.~Nickel, and M.~Le.
\newblock Flow matching for generative modeling, 2023.

\bibitem{openai2019dota2largescale}
OpenAI, :, C.~Berner, G.~Brockman, B.~Chan, V.~Cheung, P.~Dębiak, C.~Dennison, D.~Farhi, Q.~Fischer, S.~Hashme, C.~Hesse, R.~Józefowicz, S.~Gray, C.~Olsson, J.~Pachocki, M.~Petrov, H.~P. d.~O.~Pinto, J.~Raiman, T.~Salimans, J.~Schlatter, J.~Schneider, S.~Sidor, I.~Sutskever, J.~Tang, F.~Wolski, and S.~Zhang.
\newblock Dota 2 with large scale deep reinforcement learning, 2019.

\bibitem{ouyang2022traininglanguagemodelsfollow}
L.~Ouyang, J.~Wu, X.~Jiang, D.~Almeida, C.~L. Wainwright, P.~Mishkin, C.~Zhang, S.~Agarwal, K.~Slama, A.~Ray, J.~Schulman, J.~Hilton, F.~Kelton, L.~Miller, M.~Simens, A.~Askell, P.~Welinder, P.~Christiano, J.~Leike, and R.~Lowe.
\newblock Training language models to follow instructions with human feedback, 2022.

\bibitem{rombach2022highresolutionimagesynthesislatent}
R.~Rombach, A.~Blattmann, D.~Lorenz, P.~Esser, and B.~Ommer.
\newblock High-resolution image synthesis with latent diffusion models, 2022.

\bibitem{schulman2015trust}
J.~Schulman, S.~Levine, P.~Abbeel, M.~Jordan, and P.~Moritz.
\newblock Trust region policy optimization.
\newblock In {\em International conference on machine learning}, pages 1889--1897. PMLR, 2015.

\bibitem{schulman2018highdimensionalcontinuouscontrolusing}
J.~Schulman, P.~Moritz, S.~Levine, M.~Jordan, and P.~Abbeel.
\newblock High-dimensional continuous control using generalized advantage estimation, 2018.

\bibitem{schulman2017proximalpolicyoptimizationalgorithms}
J.~Schulman, F.~Wolski, P.~Dhariwal, A.~Radford, and O.~Klimov.
\newblock Proximal policy optimization algorithms, 2017.

\bibitem{sohl2015deep}
J.~Sohl-Dickstein, E.~Weiss, N.~Maheswaranathan, and S.~Ganguli.
\newblock Deep unsupervised learning using nonequilibrium thermodynamics.
\newblock In {\em International conference on machine learning}, pages 2256--2265. pmlr, 2015.

\bibitem{song2019generative}
Y.~Song and S.~Ermon.
\newblock Generative modeling by estimating gradients of the data distribution.
\newblock {\em Advances in neural information processing systems}, 32, 2019.

\bibitem{song2020score}
Y.~Song, J.~Sohl-Dickstein, D.~P. Kingma, A.~Kumar, S.~Ermon, and B.~Poole.
\newblock Score-based generative modeling through stochastic differential equations.
\newblock {\em arXiv preprint arXiv:2011.13456}, 2020.

\bibitem{sutton1999policy}
R.~S. Sutton, D.~McAllester, S.~Singh, and Y.~Mansour.
\newblock Policy gradient methods for reinforcement learning with function approximation.
\newblock {\em Advances in neural information processing systems}, 12, 1999.

\bibitem{todorov2012mujoco}
E.~Todorov, T.~Erez, and Y.~Tassa.
\newblock Mujoco: A physics engine for model-based control.
\newblock In {\em 2012 IEEE/RSJ International Conference on Intelligent Robots and Systems}, pages 5026--5033. IEEE, 2012.

\end{thebibliography}


\appendix
\section{Derivations Problem of CFG in Policy}
\label{appendix:cfg}

\paragraph{Derivation of CFG.}
In the diffusion model, the key goal is to approximate a score function $\nabla_x \log p(x|y)$. The key distribution to estimate is $ p(x|y)$ where as $y$ is the conditional label and $x$ is the target data which could be images. $p$ is the distribution parameterized by $\theta$ we are approximating to. Following \cite{dieleman2022guidance}, we can derive the CFG with Bayes’ rule:
\begin{equation*}
  \begin{aligned}
p(x|y) &= \frac{p(y|x)p(x)}{p(y)}, \\
\log p(x|y) &= \log p(y|x) + \log p(x) - \log p(y).
  \end{aligned}
\end{equation*}
Taking the derivative with respect to $x$ eliminates the $\log p(y)$ term:
\begin{equation*}
\nabla_x \log p(x|y) = \nabla_x \log p(y|x) + \nabla_x \log p(x).
\end{equation*}
Introducing a scaling parameter $\gamma$ for the conditional term we can get the classifier guidance:
\begin{equation}
\nabla_x \log \hat{p}(x|y) = \gamma \nabla_x \log p(y|x) + \nabla_x \log p(x).
\label{eq:classifier_guidance}
\end{equation}
Applying Bayes’ rule again to $p(y|x)$ gives
\begin{equation*}
\nabla_x \log p(y|x) = \nabla_x \log p(x|y) - \nabla_x \log p(x),
\end{equation*}
which simplifies Equation \ref{eq:classifier_guidance} to
\begin{equation}
\nabla_x \log \hat{p}(x|y) = (1-\gamma)\nabla_x \log p(x) + \gamma \nabla_x \log p(x|y).
\end{equation}

However, to see more clearly what it does, we can revert the derivative and exponentiate both sides which yields the reweighted distribution
\begin{equation}
\hat{p}(x|y) \propto p(x)^{1-\gamma} p(x|y)^\gamma.
\end{equation}

\paragraph{Failure of Direct Analogy in Policy Gradient.}
If we attempt the same trick with policies, Bayes’ rule gives
\begin{equation*}
\pi_\theta(a|s) = \frac{\pi_\theta(s|a)\pi_\theta(a)}{\pi_\theta(s)},
\end{equation*}
and thus
\begin{equation*}
\log \pi_\theta(a|s) = \log \pi_\theta(s|a) + \log \pi_\theta(a) - \log \pi_\theta(s).
\end{equation*}
Differentiating with respect to $\theta$:
\begin{equation*}
\nabla_\theta \log \pi_\theta(a|s) = \nabla_\theta \log \pi_\theta(s|a) + \nabla_\theta \log \pi_\theta(a) - \nabla_\theta \log \pi_\theta(s).
\end{equation*}
The last term, $\nabla_\theta \log \pi_\theta(s)$, depends on the marginal state distribution induced by both the policy and the environment. Moreover, it causes a problem of interpretation, what would the state input into policy be? Unlike in the diffusion case, it cannot be eliminated. This prevents a direct application of the CFG trick to standard policy gradients.

\section{Detailed Results}
\label{sec:detailed_results}
\begin{table}[h!]
\small
\centering
\caption{Discrete Environment Results: Mean ± 95\% CI. PPO w/ $\gamma$ is trained with $\gamma=1.1$. We report the results of the same model with different $\gamma$ during inference. }
\label{tab:discrete_results}
\begin{tabular}{lcc}
\toprule
Method & Acrobot-v1 & CartPole-v1 \\
\midrule
\multicolumn{3}{l}{\textbf{100k Steps}} \\
PPO & $-111.3 \pm 7.5$ & $270.6 \pm 81.3$ \\
\addlinespace
PPO w/ $\gamma$ & & \\
\quad $\gamma=1$ & $-111.7 \pm 7.0$ & $337.8 \pm 96.4$ \\
\quad $\gamma=1.5$ & $-102.0 \pm 6.2$ & $385.9 \pm 136.0$ \\
\quad $\gamma=2$ & $-93.7 \pm 10.2$ & $393.0 \pm 132.1$ \\
\quad $\gamma=5$ & $-87.4 \pm 3.4$ & $420.6 \pm 104.2$ \\
\quad $\gamma=10$ & $-84.0 \pm 4.8$ & $444.8 \pm 85.3$ \\
\quad $\gamma=20$ & $-82.7 \pm 4.4$ & $458.6 \pm 66.2$ \\
\midrule
\multicolumn{3}{l}{\textbf{200k Steps}} \\
PPO & $-85.6 \pm 3.1$ & $456.6 \pm 50.8$ \\
\addlinespace
\addlinespace
PPO w/ $\gamma$ & & \\
\quad $\gamma=1$ & $-86.8 \pm 6.0$ & $466.8 \pm 27.5$ \\
\quad $\gamma=1.5$ & $-87.1 \pm 2.5$ & $498.9 \pm 2.1$ \\
\quad $\gamma=2$ & $-83.5 \pm 3.5$ & $500.0 \pm 0.0$ \\
\quad $\gamma=5$ & $-83.8 \pm 2.9$ & $500.0 \pm 0.0$ \\
\quad $\gamma=10$ & $-82.8 \pm 1.3$ & $500.0 \pm 0.0$ \\
\quad $\gamma=20$ & $-82.5 \pm 1.9$ & $500.0 \pm 0.0$ \\
\midrule
\multicolumn{3}{l}{\textbf{300k Steps}} \\
PPO & $-84.7 \pm 2.3$ & $483.8 \pm 20.8$ \\
\addlinespace
\addlinespace
PPO w/ $\gamma$ & & \\
\quad $\gamma=1$ & $-85.4 \pm 6.5$ & $458.0 \pm 37.7$ \\
\quad $\gamma=1.5$ & $-80.8 \pm 4.0$ & $499.9 \pm 0.4$ \\
\quad $\gamma=2$ & $-81.7 \pm 2.5$ & $499.7 \pm 0.8$ \\
\quad $\gamma=5$ & $-82.2 \pm 0.5$ & $500.0 \pm 0.0$ \\
\quad $\gamma=10$ & $-82.6 \pm 2.4$ & $500.0 \pm 0.0$ \\
\quad $\gamma=20$ & $-83.0 \pm 1.2$ & $500.0 \pm 0.0$ \\
\midrule
\multicolumn{3}{l}{\textbf{400k Steps}} \\
PPO & $-82.4 \pm 2.4$ & $487.2 \pm 7.8$ \\
\addlinespace
\addlinespace
PPO w/ $\gamma$ & & \\
\quad $\gamma=1$ & $-82.2 \pm 3.5$ & $465.8 \pm 17.6$ \\
\quad $\gamma=1.5$ & $-81.6 \pm 2.0$ & $500.0 \pm 0.0$ \\
\quad $\gamma=2$ & $-82.7 \pm 2.9$ & $500.0 \pm 0.0$ \\
\quad $\gamma=5$ & $-83.8 \pm 2.6$ & $500.0 \pm 0.0$ \\
\quad $\gamma=10$ & $-83.0 \pm 3.0$ & $500.0 \pm 0.0$ \\
\quad $\gamma=20$ & $-82.1 \pm 2.3$ & $500.0 \pm 0.0$ \\
\midrule
\multicolumn{3}{l}{\textbf{500k Steps}} \\
PPO & $-83.6 \pm 3.2$ & $487.8 \pm 6.4$ \\
\addlinespace
\addlinespace
PPO w/ $\gamma$ & & \\
\quad $\gamma=1$ & $-82.8 \pm 1.9$ & $473.2 \pm 16.5$ \\
\quad $\gamma=1.5$ & $-83.0 \pm 2.9$ & $499.8 \pm 0.6$ \\
\quad $\gamma=2$ & $-82.9 \pm 2.8$ & $500.0 \pm 0.0$ \\
\quad $\gamma=5$ & $-82.7 \pm 2.1$ & $500.0 \pm 0.0$ \\
\quad $\gamma=10$ & $-82.5 \pm 3.5$ & $500.0 \pm 0.0$ \\
\quad $\gamma=20$ & $-82.5 \pm 3.6$ & $500.0 \pm 0.0$ \\
\bottomrule
\end{tabular}
\end{table}

\begin{table*}[htbp]
\tiny
\centering
\caption{Continuous Environment Results: Mean ± 95\% CI. PPO w/ $\gamma$ is trained with $\gamma=1.1$. We report the results of the same model with different $\gamma$ during inference.}
\label{tab:continuous_results}
\begin{tabular}{l|cccccc}
\toprule
\textbf{Method} & \textbf{HalfCheetah-v4} & \textbf{Hopper-v4} & \textbf{Humanoid-v4} & \textbf{InvertedPendulum-v4} & \textbf{Pusher-v4} & \textbf{Walker2d-v4} \\
\midrule
\multicolumn{7}{c}{\textbf{200K Steps}} \\
\midrule
PPO & $987.0 \pm 266.7$ & $1279.7 \pm 171.6$ & $456.9 \pm 28.7$ & $845.2 \pm 120.1$ & $-82.0 \pm 6.6$ & $542.0 \pm 184.3$ \\
\midrule
\multicolumn{7}{l}{\textbf{PPO w/ $\gamma$}} \\
$\gamma=1.0$ & $980.1 \pm 232.7$ & $1161.0 \pm 197.1$ & $447.4 \pm 15.7$ & $754.0 \pm 131.4$ & $-76.2 \pm 8.2$ & $749.2 \pm 272.3$ \\
$\gamma=1.05$ & $990.8 \pm 218.3$ & $1315.0 \pm 262.5$ & $475.6 \pm 27.5$ & $781.5 \pm 148.9$ & $-76.8 \pm 7.7$ & $833.2 \pm 361.1$ \\
$\gamma=1.1$ & $1022.9 \pm 244.1$ & $1392.8 \pm 401.3$ & $457.2 \pm 15.2$ & $879.9 \pm 85.9$ & $-78.0 \pm 8.4$ & $821.2 \pm 321.8$ \\
$\gamma=1.15$ & $990.5 \pm 215.9$ & $1479.5 \pm 429.6$ & $476.1 \pm 27.2$ & $921.4 \pm 63.8$ & $-78.5 \pm 8.7$ & $932.3 \pm 442.9$ \\
$\gamma=1.2$ & $1019.7 \pm 271.1$ & $1489.3 \pm 594.6$ & $474.2 \pm 26.1$ & $950.2 \pm 45.0$ & $-79.1 \pm 8.8$ & $933.2 \pm 391.3$ \\
$\gamma=1.25$ & $989.7 \pm 217.2$ & $1564.4 \pm 677.5$ & $459.3 \pm 19.0$ & $932.7 \pm 76.5$ & $-80.9 \pm 9.3$ & $901.6 \pm 355.2$ \\
$\gamma=1.3$ & $979.2 \pm 286.5$ & $1592.8 \pm 606.4$ & $467.1 \pm 17.1$ & $966.9 \pm 39.3$ & $-81.3 \pm 9.3$ & $1012.1 \pm 472.5$ \\
$\gamma=1.5$ & $923.2 \pm 354.4$ & $1763.3 \pm 868.5$ & $426.8 \pm 29.1$ & $992.4 \pm 10.4$ & $-86.0 \pm 10.9$ & $956.9 \pm 488.4$ \\
\midrule
\multicolumn{7}{c}{\textbf{400K Steps}} \\
\midrule
PPO & $1419.3 \pm 422.1$ & $1916.3 \pm 315.2$ & $539.7 \pm 31.4$ & $884.3 \pm 132.4$ & $-71.1 \pm 13.4$ & $1579.1 \pm 850.9$ \\
\midrule
\multicolumn{7}{l}{\textbf{PPO w/ $\gamma$}} \\
$\gamma=1.0$ & $1684.8 \pm 838.2$ & $1368.9 \pm 280.1$ & $503.3 \pm 33.9$ & $683.5 \pm 273.5$ & $-60.0 \pm 8.7$ & $1315.2 \pm 281.3$ \\
$\gamma=1.05$ & $1748.4 \pm 915.2$ & $1430.0 \pm 239.1$ & $535.3 \pm 40.9$ & $749.4 \pm 265.3$ & $-60.7 \pm 9.0$ & $1715.3 \pm 715.0$ \\
$\gamma=1.1$ & $1678.8 \pm 775.3$ & $1441.5 \pm 369.2$ & $558.6 \pm 40.7$ & $833.5 \pm 211.7$ & $-61.3 \pm 9.6$ & $2047.7 \pm 789.2$ \\
$\gamma=1.15$ & $1704.1 \pm 841.2$ & $1556.9 \pm 448.8$ & $561.1 \pm 37.2$ & $878.2 \pm 187.2$ & $-62.7 \pm 11.3$ & $2146.8 \pm 776.2$ \\
$\gamma=1.2$ & $1640.2 \pm 874.9$ & $1638.5 \pm 601.9$ & $557.3 \pm 33.3$ & $925.7 \pm 89.0$ & $-62.6 \pm 10.3$ & $2212.8 \pm 666.7$ \\
$\gamma=1.25$ & $1610.7 \pm 811.9$ & $1690.2 \pm 934.7$ & $542.6 \pm 23.9$ & $938.2 \pm 85.8$ & $-62.8 \pm 10.0$ & $2194.3 \pm 639.7$ \\
$\gamma=1.3$ & $1561.4 \pm 782.2$ & $1797.9 \pm 871.3$ & $526.5 \pm 30.1$ & $941.1 \pm 88.3$ & $-63.6 \pm 10.9$ & $2063.6 \pm 687.8$ \\
$\gamma=1.5$ & $1468.6 \pm 937.9$ & $1945.7 \pm 1180.0$ & $463.9 \pm 25.5$ & $982.0 \pm 38.0$ & $-67.9 \pm 14.1$ & $1618.3 \pm 566.6$ \\
\midrule
\multicolumn{7}{c}{\textbf{600K Steps}} \\
\midrule
PPO & $1654.5 \pm 746.3$ & $1857.6 \pm 237.7$ & $592.4 \pm 62.6$ & $882.0 \pm 113.1$ & $-61.0 \pm 13.3$ & $2461.5 \pm 651.1$ \\
\midrule
\multicolumn{7}{l}{\textbf{PPO w/ $\gamma$}} \\
$\gamma=1.0$ & $1998.5 \pm 1138.8$ & $1772.5 \pm 368.2$ & $545.7 \pm 46.3$ & $787.0 \pm 107.5$ & $-54.6 \pm 8.0$ & $1700.6 \pm 675.2$ \\
$\gamma=1.05$ & $1983.4 \pm 1079.0$ & $1994.3 \pm 464.1$ & $582.4 \pm 57.7$ & $882.9 \pm 86.9$ & $-54.0 \pm 5.5$ & $2150.6 \pm 636.2$ \\
$\gamma=1.1$ & $2035.1 \pm 1151.2$ & $2091.1 \pm 574.7$ & $604.5 \pm 30.9$ & $920.9 \pm 84.1$ & $-54.5 \pm 5.5$ & $2380.8 \pm 624.9$ \\
$\gamma=1.15$ & $2032.9 \pm 1145.1$ & $2146.4 \pm 374.8$ & $609.0 \pm 57.9$ & $941.8 \pm 51.2$ & $-55.4 \pm 5.3$ & $2718.5 \pm 727.1$ \\
$\gamma=1.2$ & $1940.9 \pm 1042.4$ & $2192.1 \pm 418.7$ & $580.6 \pm 49.6$ & $944.7 \pm 47.9$ & $-55.8 \pm 6.4$ & $3001.5 \pm 1028.5$ \\
$\gamma=1.25$ & $1877.3 \pm 1018.1$ & $2287.5 \pm 455.0$ & $560.2 \pm 49.1$ & $974.2 \pm 22.5$ & $-56.8 \pm 4.2$ & $3006.4 \pm 847.9$ \\
$\gamma=1.3$ & $1882.1 \pm 1055.4$ & $2248.1 \pm 613.1$ & $560.2 \pm 36.8$ & $980.9 \pm 19.5$ & $-57.2 \pm 4.8$ & $2737.1 \pm 963.6$ \\
$\gamma=1.5$ & $1678.6 \pm 1005.3$ & $2363.0 \pm 708.2$ & $489.4 \pm 35.9$ & $988.6 \pm 18.0$ & $-60.2 \pm 5.2$ & $2109.7 \pm 1193.1$ \\
\midrule
\multicolumn{7}{c}{\textbf{800K Steps}} \\
\midrule
PPO & $1792.0 \pm 941.0$ & $2337.7 \pm 603.4$ & $651.1 \pm 74.8$ & $825.5 \pm 191.1$ & $-56.8 \pm 13.4$ & $3297.3 \pm 1053.7$ \\
\midrule
\multicolumn{7}{l}{\textbf{PPO w/ $\gamma$}} \\
$\gamma=1.0$ & $2150.1 \pm 1293.8$ & $1973.6 \pm 353.0$ & $596.4 \pm 53.1$ & $855.1 \pm 196.5$ & $-51.4 \pm 8.0$ & $2152.4 \pm 1017.1$ \\
$\gamma=1.05$ & $2166.3 \pm 1289.1$ & $2113.9 \pm 369.4$ & $640.0 \pm 53.9$ & $886.8 \pm 181.9$ & $-51.3 \pm 8.7$ & $2765.3 \pm 556.1$ \\
$\gamma=1.1$ & $2090.4 \pm 1126.8$ & $2330.5 \pm 426.7$ & $635.4 \pm 55.3$ & $914.4 \pm 175.0$ & $-51.9 \pm 9.1$ & $3219.7 \pm 561.3$ \\
$\gamma=1.15$ & $2105.1 \pm 1157.5$ & $2317.8 \pm 569.8$ & $649.6 \pm 44.3$ & $913.7 \pm 201.2$ & $-52.0 \pm 9.0$ & $3245.5 \pm 529.2$ \\
$\gamma=1.2$ & $1990.0 \pm 1012.8$ & $2454.3 \pm 506.1$ & $625.9 \pm 44.0$ & $944.0 \pm 107.6$ & $-52.4 \pm 8.6$ & $3345.5 \pm 729.3$ \\
$\gamma=1.25$ & $1950.0 \pm 1003.0$ & $2517.1 \pm 670.0$ & $588.7 \pm 39.3$ & $975.9 \pm 39.4$ & $-53.7 \pm 6.4$ & $3034.7 \pm 879.8$ \\
$\gamma=1.3$ & $1929.9 \pm 1012.5$ & $2571.3 \pm 631.0$ & $571.2 \pm 51.2$ & $987.6 \pm 21.5$ & $-53.9 \pm 6.1$ & $3129.2 \pm 1074.8$ \\
$\gamma=1.5$ & $1641.6 \pm 842.1$ & $2675.4 \pm 591.1$ & $499.7 \pm 66.3$ & $991.6 \pm 23.4$ & $-56.5 \pm 6.1$ & $2371.5 \pm 1098.8$ \\
\midrule
\multicolumn{7}{c}{\textbf{1M Steps}} \\
\midrule
PPO & $1846.9 \pm 1026.3$ & $2143.1 \pm 309.7$ & $660.7 \pm 96.2$ & $864.7 \pm 143.8$ & $-55.8 \pm 12.3$ & $3056.0 \pm 950.5$ \\
\midrule
\multicolumn{7}{l}{\textbf{PPO w/ $\gamma$}} \\
$\gamma=1.0$ & $2220.4 \pm 1367.1$ & $1604.4 \pm 527.1$ & $613.6 \pm 52.9$ & $835.7 \pm 104.7$ & $-48.1 \pm 8.3$ & $2295.6 \pm 1334.7$ \\
$\gamma=1.05$ & $2257.0 \pm 1411.6$ & $1752.1 \pm 684.3$ & $677.2 \pm 29.2$ & $896.5 \pm 108.8$ & $-48.6 \pm 7.9$ & $2673.0 \pm 918.1$ \\
$\gamma=1.1$ & $2189.3 \pm 1275.9$ & $1880.3 \pm 813.9$ & $668.8 \pm 51.6$ & $905.7 \pm 88.3$ & $-49.5 \pm 7.5$ & $3269.9 \pm 824.8$ \\
$\gamma=1.15$ & $2099.4 \pm 1175.1$ & $1883.3 \pm 840.3$ & $675.2 \pm 69.8$ & $932.7 \pm 62.4$ & $-50.1 \pm 7.3$ & $3399.7 \pm 650.0$ \\
$\gamma=1.2$ & $2108.3 \pm 1184.2$ & $2048.6 \pm 1045.7$ & $656.5 \pm 80.5$ & $959.4 \pm 68.8$ & $-49.9 \pm 7.2$ & $3508.7 \pm 694.3$ \\
$\gamma=1.25$ & $2038.1 \pm 1119.5$ & $2026.3 \pm 1066.5$ & $634.0 \pm 72.0$ & $974.6 \pm 25.3$ & $-50.9 \pm 7.3$ & $3339.5 \pm 620.2$ \\
$\gamma=1.3$ & $1958.0 \pm 1042.8$ & $2162.3 \pm 1122.8$ & $587.4 \pm 62.9$ & $989.5 \pm 13.7$ & $-52.1 \pm 6.8$ & $2819.6 \pm 1227.4$ \\
$\gamma=1.5$ & $1650.4 \pm 929.2$ & $1974.8 \pm 1132.2$ & $511.8 \pm 81.2$ & $995.0 \pm 9.8$ & $-55.3 \pm 6.9$ & $2253.7 \pm 1765.3$ \\
\bottomrule
\end{tabular}
\end{table*}

\end{document}